\documentclass[letterpaper, 10 pt, conference]{ieeeconf}  

\IEEEoverridecommandlockouts                              

\usepackage{amsmath}
\usepackage{hyperref}
\usepackage{accents}
\usepackage{algorithm}  
\usepackage{algorithmic}
\usepackage{url}
\usepackage{svg}
\usepackage{multirow}
\usepackage{subfiles}
\usepackage{biblatex}
\usepackage{subcaption}
\usepackage{makecell}

\iffalse
    \newcommand{\holger}[1]{\noindent}
\else
    \newcommand{\holger}[1]{{\color{blue}[HC: #1]}} 
\fi

\graphicspath{{figures/}}
\title{\LARGE \bf
Offline Tracking with Object Permanence
}

\author{Xianzhong Liu$^{1}$ and  Holger Caesar$^{1}$
\thanks{$^{a}$ Intelligent Vehicles section, Delft University of Technology, The Netherlands}
}

\begin{document}

\maketitle
\thispagestyle{empty}
\pagestyle{empty}

\begin{abstract}
To reduce the expensive labor costs of manually labeling autonomous driving datasets, an alternative is to automatically label the datasets using an offline perception system. However, objects might be temporarily occluded. Such occlusion scenarios in the datasets are common yet underexplored in offline auto labeling. In this work, we propose an offline tracking model that focuses on occluded object tracks. It leverages the concept of object permanence, which means objects continue to exist even if they are not observed anymore. The model contains three parts: a standard online tracker, a re-identification (Re-ID) module that associates tracklets before and after occlusion, and a track completion module that completes the fragmented tracks. The Re-ID module and the track completion module use the vectorized lane map as a prior to refine the tracking results with occlusion. The model can effectively recover the occluded object trajectories. It significantly improves the original online tracking result, demonstrating its potential to be applied in offline auto labeling as a useful plugin to improve tracking by recovering occlusions.
\end{abstract}


\section{INTRODUCTION}
Supervised deep learning-based models have achieved good performance in autonomous driving. However, it usually requires a huge amount of labeled data with high quality to train and tune such data-hungry models. 
An effective way is to auto label datasets, where labels can be automatically provided by a trained perception system. Waymo first proposed to auto label data offline to improve the quality of the generated labels~\cite{offboard_labeling}. In online tracking, the location of an object is inferred only from past and present sensor data. Online trackers are thus likely to produce false associations under severe occlusions. Offline multi-object tracking (MOT) is acausal and the position of an object can be inferred from past, present, and future sensor data. A consistent estimate of the scene can thus be optimized globally using the data not limited to a short moment in the past, enabling accurate object tracking even under severe occlusions. Based on global information,~\cite{Auto4d,offboard_labeling, CTRL, Detzero} have developed offline auto labeling pipelines that generate accurate object trajectories in 3D space from LiDAR point cloud sequence data.

 Previous offline labeling methods mainly focused on using the observation sequence to refine the unoccluded trajectories (e.g. location, orientation, and size of the bounding boxes). However, object trajectories and point cloud sequences are sometimes partially missing because of the temporary occlusions on the objects. Severely occluded objects are likely to be missed by detectors and thus cannot be continuously tracked. Furthermore, the trackers in~\cite{Auto4d,offboard_labeling} adopt a death memory strategy to terminate the unmatched tracks, which means that a track not matched with any detections for a predefined number of frames will be terminated. Therefore, the identity of the same object may switch after reappearance from occlusion. In~\cite{Object_permanence}, the tracker learns object permanence from synthetic data by hallucinating the occluded motion. 
 The authors use pseudo-labels with constant velocity to supervise the full occlusion scenarios, which fails to capture the nonlinear motion behind occlusions. The direct supervision using ground truth (GT) labels yields a worse performance than using the constant-velocity pseudo-labels. Without any extra prior knowledge, a neural network is not able to learn the complex motion and instead resorts to using a linear velocity baseline.
 Immortal Tracker~\cite{Immortaltrackers} also captures object permanence by not terminating any unmatched tracks and prolonging their predictions. It has been applied in~\cite{CTRL, Detzero} for its effectiveness in reducing identity switches (IDS). However, the data association is still simply done by using the constant-velocity prediction from a Kalman filter~\cite{kalman_filter} which cannot capture the nonlinear motion under long occlusions either. Thus, tracking under long occlusion remains a challenge in point cloud-based offline auto labeling. 
 Motion prediction models~\cite{laneGCN, gao2020vectornet}, on the other hand, can produce accurate vehicle trajectories over longer horizons based on a semantic map. The lanes on the map serve as a strong prior knowledge to guide the motion of target vehicles and thus can be used to estimate motion under occlusion. Based on this insight, we not only associate tracklets together accurately but also complete the fragmented tracks. Unlike Immortal Tracker~\cite{Immortaltrackers}, we additionally interpolate and refine the occluded trajectory based on the lane map feature.

In this work, we propose an offline tracking model containing three modules, as shown in Fig.~\ref{fig: pipeline} to tackle the aforementioned problems. The first module is an off-the-shelf online tracker which produces the initial tracking result. Leveraging the idea of object permanence, the second Re-ID module tries to reassociate the terminated tracklets with the possible future candidate tracklets. 
Based on the association result, the last track completion module completes the missing trajectories. Taking the insight from the motion prediction task, we have extracted map information as a prior to enhance the association and track completion modules. Unlike common motion prediction methods with a fixed prediction horizon predefined~\cite{PGP, HOME}, our model can use a flexible prediction horizon at inference time to decode the predicted poses in the track completion module, to deal with variable occlusion durations. Compared to goal-based motion prediction methods~\cite{HOME}, we take future tracklets as inputs instead of goal points, which enables the model to predict more accurately based on motion patterns.
%
Previous offline auto-labeling methods primarily prioritize the precision of visible bounding boxes as their main metric, with MOT metrics receiving minor attention. In contrast, our work focuses on optimizing and evaluating MOT metrics.

The contributions of this paper are summarized as follows:
\vspace{-2mm}
\begin{itemize}
\item We propose an offline tracking model for Re-ID (Sect.~\ref{sec: Re-ID}) and occluded track completion (Sect.~\ref{sec: track completion}) to track occluded vehicles.
\item We are the first to utilize vectorized lane maps in the MOT task. By using these maps as a prior, we are able to recover the complex motion under occlusion.
\item By encoding variable time queries (Sect.~\ref{sec: track completion}), we present the first method that can decode future trajectories of variable duration, accommodating occlusions of differing lengths.
\item We optimize our method for the MOT task and demonstrate the improvements relative to the original online tracking results on the nuScenes dataset (Sect.~\ref{sec: online tracking eval}), showing the potential of our model to be applied in offline auto labeling as a useful plugin to improve tracking and recover occlusions.
\item The code and pre-trained models are publicly available at \href{https://github.com/tudelft-iv-students/offline-tracking-with-object-permanence.git}{github.com/tudelft-iv-students/offline-tracking-with-object-permanence}.
\end{itemize}    

\vspace{-1mm}
\section{RELATED WORK}

\label{related_work}
\vspace{-1mm}
\subsection{Tracking under occlusion}
\vspace{-1mm}
Estimating a target's motion under occlusion is one of the main challenges in the MOT task.
A common method adopts a constant velocity~\cite{basline_with_AMOTA, Immortaltrackers} transition model and uses the Kalman filter~\cite{kalman_filter} to update the tracks when detections are missed due to occlusion. CenterPoint~\cite{Centerpoint} directly estimates displacement offset (or velocity) between frames. The unmatched track is updated with a constant velocity model. PermaTrack~\cite{Object_permanence} supervises the occluded motion with pseudo-labels that keep constant velocity. Those methods heavily rely on motion heuristics. Such heuristics work well for short occlusion but cannot capture the nonlinear motion under long occlusion.
Due to the large deviation error accumulated over time, the prediction of the unmatched track would not be associated with the correct detection. Most tracking models that have finite death memory~\cite{Centerpoint,basline_with_AMOTA} would terminate such unmatched tracks, even if the object is detected again later. Such premature termination would lead to IDS. Immortal Tracker~\cite{Immortaltrackers} effectively reduces such IDS by extending the life of unmatched tracks forever. 
Our model does not rely on any handcrafted motion heuristics to associate.
Given any pair of future and history tracklets, the model outputs the learned affinity scores directly.



\vspace{-2mm}
\subsection{Offline auto labeling from lidar point clouds}
\vspace{-1mm}
Manually annotating lidar point cloud data takes much effort due to the sparsity of the data and the temporal correlation of the sequence. Several works have attempted to tackle this problem by automatically labeling the dataset~\cite{Argoverse,Nuplan,waymo_large}. 
Auto4D and 3DAL~\cite{Auto4d,offboard_labeling} fully automate annotation by taking initial tracklets generated from an online tracker (i.e. AB3DMOT~\cite{basline_with_AMOTA}). Point cloud and detection features are extracted to refine the initial tracks. Given full observation, such refinement utilizes future information to perform global optimization offline. 
Though these methods can produce accurate bounding boxes, their data association is still simply done online. The global information available in the offline setting has yet to be utilized in tracking. Such methods are likely to produce IDS and inconsistent tracks under occlusion. CTRL~\cite{CTRL} uses the Immortal Tracker~\cite{Immortaltrackers} to associate fragmented tracks due to long occlusion or missing detection. Then it backtraces the tracks by extrapolation to track the missing detections. DetZero~\cite{Detzero} also uses the Immortal Tracker to perform bidirectional tracking on the time horizon and then ensemble the results with forward and reverse order tracking fusion. As a result, CTRL and DetZero have achieved better results which surpass the human annotation. However, their data association still relies on a Kalman filter with the constant-velocity model.
\vspace{-2mm}
\subsection{Map-based prediction}
\vspace{-1mm}
Semantic maps are widely used as an input in motion prediction methods. As it contains accurate scene context from the surrounding environment, including road lanes, crosswalks, etc. With such information, models can predict how target vehicles navigate in the environment over a long horizon. HOME~\cite{ HOME} rasterizes the map as multiple vector layers with distinct RGB values representing different map elements. Alternatively, VectorNet~\cite{gao2020vectornet} proposes the vectorized approach which represents curves as vectorized polylines. It uses a subgraph network to encode each polyline as a node in a fully connected global interaction graph. It achieves better performance over the CNN baseline and reduces the model size. To capture higher resolution, LaneGCN~\cite{laneGCN} uses polyline segments as map nodes. It models a sparsely connected graph following the map topology. Similarly, PGP~\cite{PGP} constrains the connected edges such that any traversed path through the graph corresponds to a legal route for a vehicle.

\vspace{-1mm}
\section{METHOD}
\label{headings}
\vspace{-1mm}

\begin{figure*}[ht]
    \centering
    \includegraphics[width=0.9\textwidth]{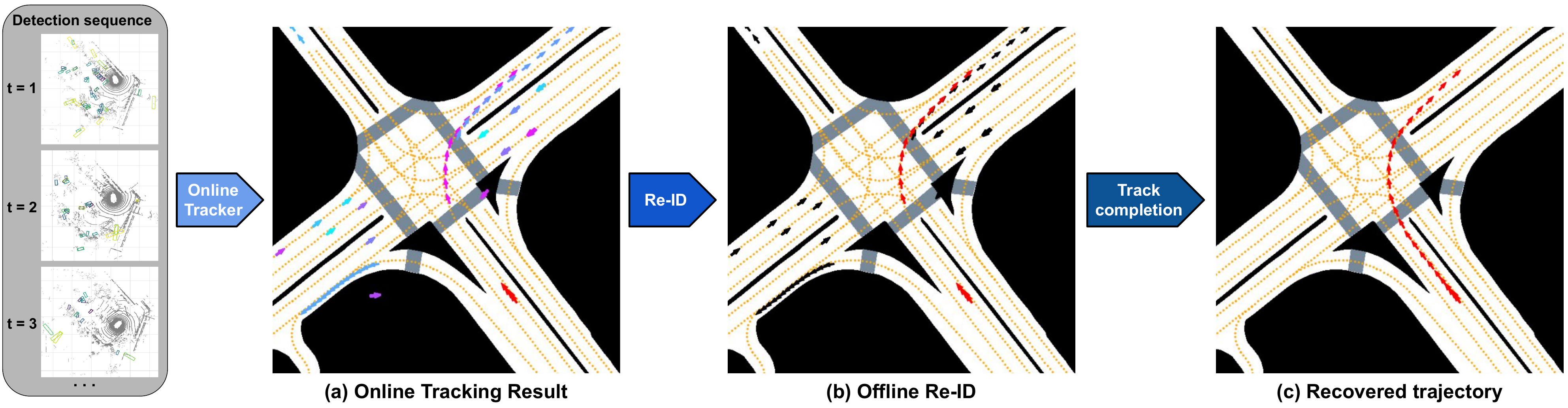}
    \caption{A brief overview of the offline tracking model. 
    \textbf{(a) Online tracking result}: 
    Each tracklet is represented by a different color (the history tracklet is red). 
    \textbf{(b) Offline Re-ID}: 
    The matched pair of tracklets are red. The unmatched ones are black. 
    \textbf{(c) Recovered trajectory}.
    }
    \vspace{-4mm}
    \label{fig: pipeline}
\end{figure*}
As shown in Fig.~\ref{fig: pipeline}, our model initially takes the detections from a detector as input. Then it uses an off-the-shelf online tracker to associate detections and generate initial tracklets. Next, the Re-ID module tries to associate the possible future tracklets with the terminated history tracklet. If a pair of tracklets are matched, the track completion module interpolates the gap between them by predicting the location and orientation of the missing box at each timestamp. Both modules extract motion information and lane map information to produce accurate results. The model finally outputs the track with refined new identities and completes the missing segments within the tracks.

\vspace{-1mm}
\subsection{Re-ID}
\label{sec: Re-ID}
\vspace{-1mm}
Based on the concept of object permanence, we assume any tracklets could potentially be terminated prematurely due to occlusion, and any other tracklets that appeared after the termination could be matched. The Re-ID model has two branches, the motion and the map branch. The motion branch takes tracklet motion as input and outputs affinity scores based on vehicle dynamics. The map branch takes motion and the lane map as inputs and outputs affinity scores based on map connectivity. Thus, for each target history tracklet, the model computes motion affinity and map-based affinity scores for it with all the possible future tracklets. The problem can thus be treated as a \textbf{binary classification task} for each single matching pair. The final score is a weighted sum of the motion and map affinity scores. We perform bipartite matching by greedily associating history tracklets with their future candidate tracklets based on the final scores. The matching pair which has a tracking score lower than a threshold is excluded from the association. 

\subsubsection{Re-ID formulation}
\label{sec: Re-ID formulation}
The upstream online tracker provides a set of tracklets $\mathbf{T}$ in the scene. In $\mathbf{T}$, history tracklets are the tracklets that end earlier before the last frame of the scene. They are represented as $\mathbf{H}=\left[H_{1}, \ldots, H_{n}\right]$. For the $i$-th history tracklet $H_{i}$, it contains history observations of $\left[ h^{i}_{-T_i}, \ldots, h^{i}_{0}\right]$, where $h^{i}_{0}$ is the last observation of $H_{i}$ on the temporal horizon and $h^{i}_{-T_i}$ the first observation. Each observation contains the location, orientation, size, uncertainty, and velocity. It is defined as:
\begin{equation}\label{eq:observation formulation}
\begin{aligned}
    h^{i}_t=\left(x^{i}_{t}, y^{i}_{t},  \theta^{i}_{t}, l^{i}_{t}, w^{i}_{t}, \mathrm{h}^{i}_{t}, s^{i}_{t}, v_{xt}^{i}, v_{yt}^{i}\right)
\end{aligned}
\end{equation}
where $l,w,\mathrm{h}$ represent the size of the $j$-th bounding box. $s$ is the confidence score, and $v_x, v_y$ the velocities on $x,y$ directions.
For each history tracklet $H_{i}$, it has a set of possible future candidate tracklets for matching: $\mathbf{F}^{i}=\left[F^{i}_{1}, \ldots, F^{i}_{n}\right]$. Each future tracklet $F^{i}_{j}$ starts after the termination of $H_{i}$: $F^{i}_{j}=\left[ f^{i}_{t_j}, \ldots, f^{i}_{t_j+T_{j}}\right],  \text{s.t.} \ t_j>\tau \label{eq:future candidate constraints}$.
In practice, we set $\tau$ as the horizon of the death memory of the online tracker since the GT future tracklets reappearing within such period are likely to be associated by the online tracker. Tracklet pairs violating this constraint will not be considered for the association. Each tracklet has a feature dimension of $T \times 8$, where $T$ is the length of the tracklet. The 8 features correspond to $\left[x, y, \theta, t, cos(\theta), sin(\theta),v_x,v_y\right] $, where $x,y$ are the local BEV coordinates, $\theta$ the yaw angle, $t$ the time relative to the last observation of the history tracklet and $v_x,v_y$ the velocities on x and y direction.

The model also extracts information from the fully connected lane graph $\mathcal{G}=(V, E)$ where each node in $V$ is a lane section. $E$ comprises edges connecting all node pairs. Each lane node is encoded from a lanelet, which is a section of the lane with a maximal length of 20m. A lanelet is represented as several discrete lane poses with a resolution of 1m. Each lane pose contains the feature of $\left[x_{lane}, y_{lane}, \theta_{lane}, cos(\theta_{lane}), sin(\theta_{lane}), \mathcal{D},\mathcal{L}_{lane}\right]$, where $x_{lane}, y_{lane}$ are the BEV coordinates of the lane poses in the local frame, $\theta_{lane}$ the yaw angle, $\mathcal{D}$ the binary flag indicating whether the lane section ends. Following PGP~\cite{PGP}, we have also included $\mathcal{L}_{lane}$, a 2-D binary vector indicating whether the pose lies on a stop line or crosswalk. The graph has a feature with the dimension of $N_{lane}\times l\times 8$, where $N_{lane}$ is the number of lane nodes and $l$ the length of a lanelet. 

Given a history tracklet $H_{i}$ and a future tracklet $F^{i}_{j}$ and lane graph $\mathcal{G}$, the model outputs motion affinity scores and map-based affinity scores:

\begin{equation}\label{eq:affinity scores}
\begin{aligned}
    C^{i,j}_{motion}=Net_{motion}(H_{i},F^{i}_{j}) \in \left[0,1\right] \\
    C^{i,j}_{map}=Net_{map}(H_{i},F^{i}_{j},\mathcal{G})  \in \left[0,1\right] 
\end{aligned}
\end{equation}
where $Net_{motion}$ and $Net_{map}$ represent the networks of the motion branch and map branch respectively. 
The final matching score for the association $C^{i,j}$ is a weighted sum of $C^{i,j}_{motion}$ and $C^{i,j}_{map}$. Therefore, we can construct a matching score matrix $\boldsymbol{C} \in n \times N$ for tracklet association, where $N$ is the number of tracklets $\mathbf{T}$ in the scene and $n$ the number of history tracklets $\mathbf{H}$. 
Each row in $\boldsymbol{C}$ represents a history tracklet and each column represents a future tracklet. Based on the established affinity matrix, bipartite matching is formulated as a linear assignment problem such that the sum of the matching scores is maximized. 
\begin{figure*}[htb]
\centering
\includegraphics[width=\textwidth]{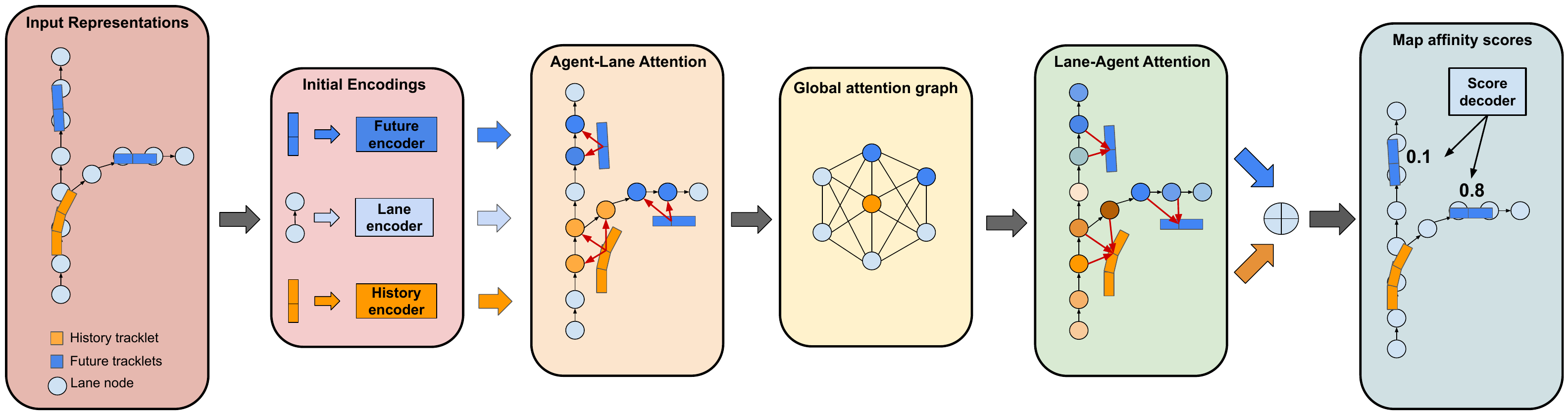}
\caption{A brief overview of the map branch. The branch starts with three parallel encoders which encode the future tracklet, lanes and history tracklet respectively. The model then propagates information between tracklets and the lane map by performing attention. Finally, map-based affinity scores are decoded from the tracklet features.}
\label{fig: map branch}
\end{figure*}
\subsubsection{Motion affinity}
\label{sec: motion affinity}

\begin{figure}
    \vspace{-4mm}
    \centering
    \includegraphics[width=0.4\textwidth]{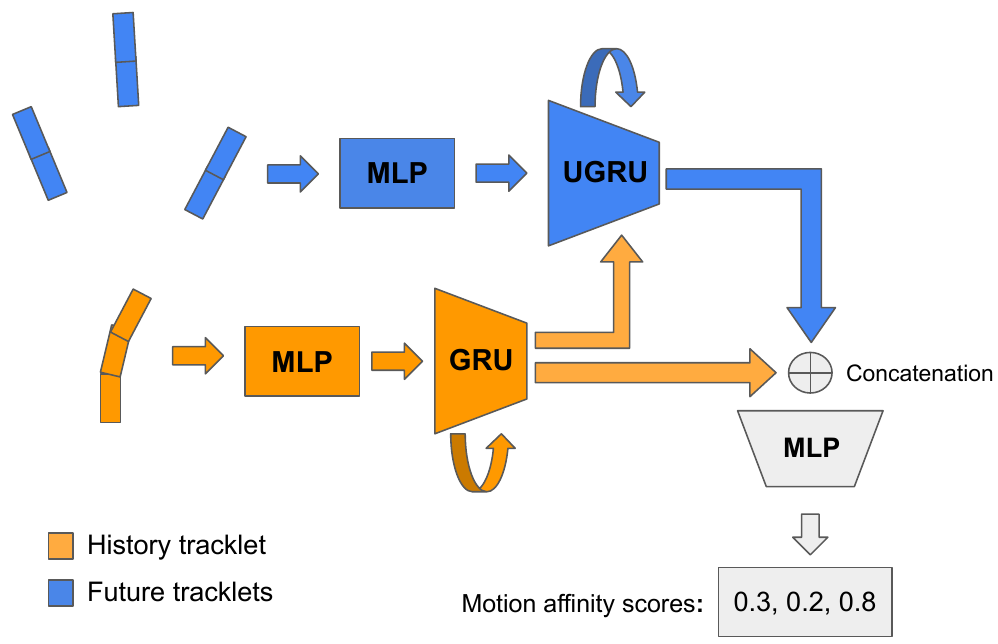}
    \caption{The network structure of motion affinity branch. The history tracklet encoder is orange, whereas the future tracklets encoder is blue. Three possible future candidates correspond to the three outputted motion affinity scores.}
    \label{fig: motion branch}
    \vspace{-5mm}
\end{figure}

We first transform the trajectories from the global frame to the local agent frame, where the origin aligns with the pose of the last observation of the history tracklet. The motion feature of each tracklet contains the location, orientation, time and velocity information at every time step. Given one history tracklet and all its possible future matching candidate tracklets, the model takes the motion features as input and outputs affinity scores in the range of $[0,1]$ for all future tracklet candidates. As in Fig.~\ref{fig: motion branch}, the motion affinity branch comprises MLPs and GRUs. The GRU encoder encodes the history motion and outputs the last hidden state as history encoding. The history encoding is then used as the initial hidden state for the UGRU~\cite{urnn} to encode future motions, enabling the awareness of historical information when encoding future motions. Unlike typical U-RNN used in motion prediction for encoding history motion features, our UGRU first does a forward pass and then does a backward pass in reverse order to encode the future motion features. The future motion encodings are concatenated with the history encoding to decode motion affinity scores. 
\subsubsection{Map-based affinity}
\label{sec: map affinity}
To improve the association accuracy, we further extract information from the lane graph as another branch. The lane centerlines provide information about the direction of traffic flow and the possible path for the vehicles to follow, which can be utilized for occluded motion estimation over a long horizon. We represent the lane map as a vectorized graph as previously described in Sect.~\ref{sec: Re-ID formulation}. Compared with the rasterized representation in~\cite{HOME}, this is a more efficient representation due to the sparsity of the map. We thus can easily extend the size of the map to incorporate all possible future tracklets far away in case of long occlusions.
As in Fig.~\ref{fig: map branch}, the map branch starts by encoding the history and the future tracklets, using the same encoder previously introduced in the motion branch (Sect.~\ref{sec: motion affinity}). The lane pose features are also encoded in parallel using a single-layer MLP. The model then performs agent-to-lane attention to fuse the information from the tracklets to the lanes. It selectively performs attention such that the tracklet encodings are only aggregated to the surrounding lanes nearby. We used the spatial attention layer in~\cite{laneGCN} to aggregate the feature from tracklets to lane poses.
The model aggregates the feature from each sub-node lane pose to get the lane node features. We apply a 2-layer-GRU to aggregate sub-node features. The first GRU layer is bidirectional so that the information can flow in both directions. 
Having the lane node features carrying the information of the tracklets, the model performs global attention~\cite{Transformer_attention} to propagate information globally in the graph.
Next, the model aggregates the updated lane features back to tracklets by applying a lane-to-agent attention layer~\cite{laneGCN}. Finally, the future and history tracklet encodings updated with global map information are concatenated together. An MLP decodes map affinity scores from the concatenated features.

\subsubsection{Association}

To match $n$ history tracklets with $N$ future tracklets, we have an affinity score matrices $\boldsymbol{C_{motion}}$ and $ \boldsymbol{C_{map}} \in n \times N$ from the two branches, we first filter the matching pairs whose affinity scores are lower than a threshold, then we get the final matching score matrix:
\begin{equation}
\boldsymbol{C}= w \cdot \boldsymbol{C_{map}} + (1-w) \cdot
 \boldsymbol{C_{motion}}
\end{equation}
where $w \in [0,1]$ is a scalar weight. During experiments, we simply set $w$ to 0.5. In practice, it can be optimized using validation data or decided based on domain knowledge about the relative reliability of motion versus map-based affinity scores. $\boldsymbol{C}$ is the weighted sum of the two affinity matrices. Finally, we solve the bipartite problem by maximizing the overall matching scores. Due to the sparsity of $\boldsymbol{C}$, we simply perform a greedy matching.
\vspace{-2mm}
\subsection{Track completion}
\vspace{-1mm}
\label{sec: track completion}

With the upstream Re-ID results, we have associated multiple tracklet pairs together. Therefore, we now have multiple fragmented tracks which have a missing segment in the middle. The track completion model interpolates the gaps within the tracks. Based on the history and future motion, the model first generates an initial trajectory in between. Then, a refinement head is applied to refine the initial trajectory based on motion and map features. The trajectory completion model learns a \textbf{regression task}. 

We first transform the future and history tracklets from the global frame to the local agent frame. We set the origin as the midpoint of the line connecting the two endpoints of the missing segment.
Most prediction models generally have a fixed prediction horizon~\cite{laneGCN,PGP}. But the occlusion horizon in reality is never fixed. We thus use time features as input queries to decode poses at all the target timesteps. Specifically, each time query feature at time $t$ is given as $\left[t, t/T \right]$, where $T$ is the total prediction horizon. So that we can have a variable number of time queries corresponding to a variable prediction horizon. 
\subsubsection{Track completion formulation}
\label{sec: Track completion formulation}
Following the formulation in Sect.~\ref{sec: Re-ID formulation}, the model receives a history observation sequence $H_{i} = \left[ h^{i}_{-T_i}, \ldots, h^{i}_{0}\right]$, and a matched future observation sequence  $F_{i}=\left[ f^{i}_{t_i}, \ldots, f^{i}_{t_i+T'_{i}}\right]$, where $t_i$ is the occlusion horizon of the $i$-th trajectory, $T_i$ and $T'_i$ the lengths of history and future sequence respectively. The model predicts a motion sequence $\mathcal{P}_i=\left[ p^{i}_{1}, \ldots, p^{i}_{t_i-1}\right]$ in between. Here, we choose to output the $x,y$ coordinates in the BEV plane, and the yaw angle $\theta$. 
\begin{equation}\label{eq:track completion formulation}
\begin{aligned}
    p^{i}_{t} = Net_{completion}(F_{i},H_{i},\mathcal{G},t)=\left[x^i_t,y^i_t,\theta^i_t \right]
\end{aligned}
\end{equation}
where $Net_{completion}$ represents the network of the track completion model. We use the same motion and map features as in the previous Sect.~\ref{sec: Re-ID formulation}, with the exception that we do not use the velocity explicitly. This results in a feature dimension of $T_{input} \times 6$. Given that the start and end pose of the predicted are already known, the model can implicitly predict the velocity in between. Hence, we discard the velocity features to reduce the possible noise from the imperfect online tracking result.  

\begin{figure}[htb]
\centering
\includegraphics[width=0.495\textwidth]{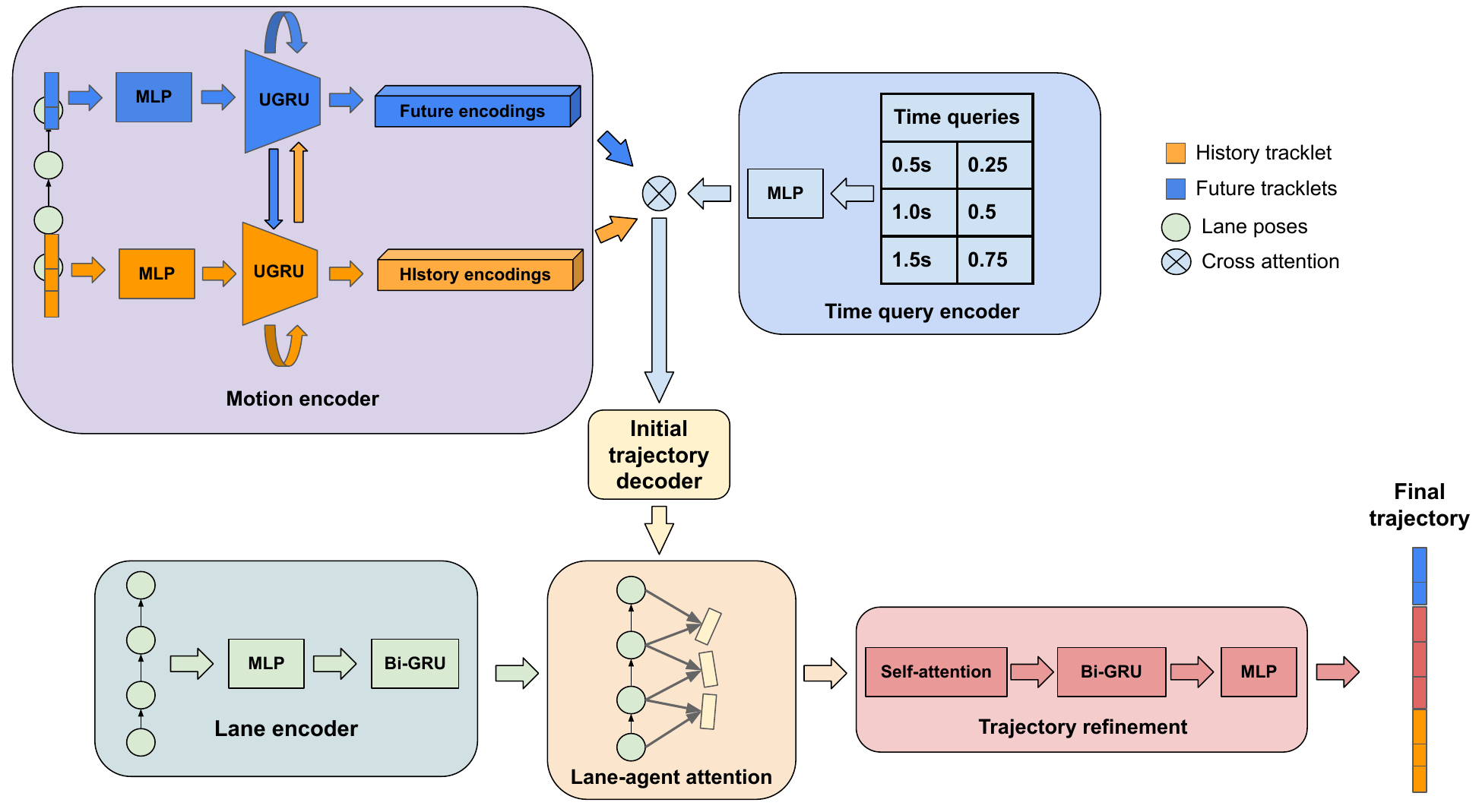}
\caption{A brief overview of the track completion model. The model stacks several sub-modules together to sequentially aggregate features and refine the results.  }
\label{fig: track completion}
\vspace{-2mm}
\end{figure}

\subsubsection{Track completion network}
\label{sec: Track completion network}
The overall structure of the track completion model is shown in  Fig.~\ref{fig: track completion}. It encodes time features as queries and iteratively aggregates context information to the query features to refine the generated tracks. The motion encoder first encodes the history and future motion features using two parallel encoders. The UGRU in each encoder takes the final hidden state from the other, hence the information can propagate through the temporal gap. The time features are encoded with a single-layer MLP as initial queries. A cross-attention layer then aggregates the motion information to the time queries. We also add a skip connection to concatenate the attention output with the motion encoding and time features. To improve the result, we added a refinement phase to the initial trajectory. The model encodes lane features and performs lane-to-agent attention~\cite{laneGCN} to fuse the extracted lane information with the generated trajectory information. The trajectory encodings carrying the lane information then go through the final trajectory refinement head which uses a self-attention layer with a skip connection to propagate information within the track. The following bidirectional GRU implicitly smooths the track by aggregating features from each pose to its adjacent poses. Finally, an MLP decodes the refined trajectory.

\vspace{-2mm}
\subsection{Loss function}
\vspace{-1mm}
We adopt the focal loss~\cite{focal_loss} to train the Re-ID model.
with $\alpha_{\mathrm{t}}$ and $\gamma$ set as $0.5$ and $2.0$ respectively. For the track completion model, we use Huber loss (i.e. smooth L1 loss) as in~\cite{laneGCN} to train the $x,y$ coordinate regression heads in both the initial trajectory decoder and the final refinement blocks.
For yaw angle regression, we first adjust all the GT yaw angles such that their absolute differences with the predicted yaw angles are less than $\pi$. 
The final regression loss $\mathcal{L}_{\text {reg }}$ is:
\begin{equation}
\label{eq: total regression loss}
\begin{aligned}
\mathcal{L}_{\text {reg }}=\alpha_{coord}\mathcal{L}_{\text {coord }} + \alpha_{yaw}\mathcal{L}_{\text {yaw }}
\end{aligned}
\end{equation}
where $\alpha_{coord}$ and $\alpha_{yaw}$ are the weights for the coordinate regression loss $\mathcal{L}_{\text {coord }}$ and yaw angle regression loss $\mathcal{L}_{\text {yaw }}$ respectively. We set $\alpha_{coord}=1.0$ and $\alpha_{yaw}=0.5$.

\vspace{-1mm}
\section{EXPERIMENT SETUP}
\vspace{-1mm}
\subsection{Training setup}
\vspace{-1mm}
\textbf{Training data:}
While~\cite{offboard_labeling,Auto4d} train their methods using online tracker outputs, we train only using the GT, but additionally inject pseudo-occlusions by masking a partial segment of each GT track.
This gives two advantages: 
\textbf{1) More accurate inputs}:
Unlike the imperfect trajectories generated from an online tracker, the trajectories from the available autonomous driving datasets are accurate and contain little noise.
With proper data augmentation, our model can still deal with imperfect data when inferencing with the trajectories produced from an online tracker.
\textbf{2) More training samples}: Since we have longer intact tracks as input, we can generate data with longer occlusions.

\textbf{Pseudo-occlusion:} 
We randomize the length of the pseudo-occlusion for each track during the training of the Re-ID and track completion models. The pseudo-occlusion duration is selected such that the history tracklet and all the future tracklets have at least one observation.

\textbf{Data augmentation:} 
To address the potential imperfections from the online tracking results, we augment the training data by randomly rotating the local frame in each sample and adding random noise to the motion inputs. 
\vspace{-1mm}


\subsection{Evaluation setup} 
\vspace{-1mm}
We train and evaluate our model on the nuScenes dataset~\cite{nuscenes}. 
The standard MOT evaluation on nuScenes could punish the fully occluded predictions since the evaluation code filters out all the GT boxes with no points inside and then linearly interpolates the GT tracks. Therefore, some of the bounding boxes recovered from occlusion could be regarded as false positives (FP) as the GT boxes they are supposed to be matched with are linearly interpolated. We thus train and validate our model on the training split of the nuScenes dataset. The evaluation is separately done on the validation split and the test split so that we can not only benchmark our method in a standard way but also adjust the evaluation procedure for occlusion cases. 
In each of the experiments below, we use one of the following three evaluation setups as indicated.

\textbf{Official nuScenes setup}: Following the official nuScenes MOT evaluation protocol, we filter out all the empty GT boxes without any lidar or radar points inside. We thus focus on the visible GT boxes. Therefore, this evaluation can also be applied to the test split on the official server.

\textbf{All-boxes setup}: To get a comprehensive evaluation result, we do not filter any GT boxes. Therefore, the evaluation takes both the visible and the occluded boxes into account.

\textbf{Pseudo-occlusion setup}: To generate more occlusions with longer duration for evaluation, we take the GT vehicle tracks from the nuScenes prediction validation set and mask a partial segment within each track. Unlike the previous setups where the model takes the online tracking result as input, the model takes the unoccluded GT tracklets as inputs under this setup. In every evaluated sample, all future tracks are masked for random durations to create diverse pseudo-occlusions. The durations range from 1.5s to each track's maximal length (12.5s at most) such that each future tracklet has at least one last pose visible. The history tracklet in each sample is also randomly deprecated such that it has at least one pose left as input and is at most 2.5 seconds long.

\vspace{-1mm}
\section{EXPERIMENT}
\vspace{-1mm}

\begin{table*}[ht!]
\centering
\caption{Comparison of MOT metrics over the top-performing methods using CenterPoint~\cite{Centerpoint} detections on the nuScenes test split (official nuScenes setup).}
\vspace{-1mm}
\begin{tabular}{c|c|c|c|c|c|c|c|c|c}
\textbf{Method} & \multicolumn{5}{c|}{\textbf{AMOTA}} & \multicolumn{1}{c|}{\textbf{AMOTP}} & \textbf{Recall$\uparrow$} & \textbf{MOTA$\uparrow$} & \textbf{IDS$\downarrow$} \\ \cline{2-6}
                & \textbf{Overall} & \textbf{Car} & \textbf{Bus} & \textbf{Truck} & \textbf{Trailer} & \textbf{$\downarrow$ \text{/ m}} & & & \\ \hline
CenterPoint~\cite{Centerpoint} & 69.8 & 82.9 & 71.1 & 59.9 & 65.1 & 0.596 & 73.5 & 59.4 & 340 \\
SimpleTrack~\cite{SimpleTrack} & 70.0 & 82.3 & 71.5 & 58.7 & 67.3 & 0.582 & 73.7 & 58.6 & 259 \\
UVTR~\cite{UVTR} & 70.1 & 83.3 & 67.2 & 58.4 & \textbf{71.6} & 0.636 & \textbf{74.6} & 59.3 & 381 \\
Immortal Tracker~\cite{Immortaltrackers} & 70.5 & 83.3 & 71.6 & 59.6 & 67.5 & 0.609 & 74.5 & 59.9 & 155 \\
NEBP~\cite{NEBP} & 70.8 & 83.5 & 70.8 & 59.8 & 69 & 0.598 & 74.1 & \textbf{61.9} & \textbf{93} \\
3DMOTFormer++~\cite{3DMOTFormer} & 72.3 & 82.1 & 74.9 & 62.6 & 69.6 & 0.542 & 73.0 & 58.6 & 210 \\
ShaSTA~\cite{ShaSTA} & 73.1 & 83.8 & 73.3 & \textbf{65} & 70.4 & 0.559 & 74.3 & 61.2 & 185 \\
\hline
Offline Re-ID (Motion + Map) & \textbf{73.4} & \textbf{84.2} & \textbf{75.1} & 64.1 & 70.3 & \textbf{0.532} & 74.2 & 61.3 & 204 \\
\end{tabular}
\vspace{-3mm}
\label{tab: merged Re-ID AMOTA and MOT metrics test}
\end{table*}

\begin{table*}[ht!]
\centering
\caption{Track completion evaluation on the nuScenes validation split (All-boxes setup). w: with Re-ID and track completion. w/o: without Re-ID and track completion. }
\label{tab: track completion results online}
\vspace{-1mm}
\begin{tabular}{c|ccc|ccc|ccc|ccc}
\multirow{2}{*}{\begin{tabular}[c]{@{}c@{}}\textbf{Track Completion}\\ { }\end{tabular}} & \multicolumn{3}{c|}{\textbf{AMOTA$\uparrow$}} & \multicolumn{3}{c|}{\textbf{AMOTP$\downarrow$ \text{/ m}}} & \multicolumn{3}{c|}{\textbf{IDS$\downarrow$}} & \multicolumn{3}{c}{\textbf{Recall$\uparrow$}} \\ \cline{2-13} 
                                                                             & w/o     & w    & $\Delta$            & w/o     & w  & $\Delta$              & w/o    & w     & $\Delta$          & w/o     & w    & $\Delta$            \\ \hline
CenterPoint~\cite{Centerpoint}                                                                  & 70.2    & \textbf{72.4}  & \textcolor{green}{\textbf{+2.2}}  & 0.634   & \textbf{0.615} & \textcolor{green}{\textbf{-0.019}}  & 254    & \textbf{183}  & \textcolor{green}{\textbf{-71}}  & 73.7    & \textbf{74.5}  & \textcolor{green}{\textbf{+0.8}}  \\
SimpleTrack~\cite{SimpleTrack}                                                              & 70.0    & \textbf{71.0} & \textcolor{green}{\textbf{+1.0}}   & 0.668   & \textbf{0.629} & \textcolor{green}{\textbf{-0.039}}  & 210    & \textbf{170}  & \textcolor{green}{\textbf{-40}}  & 72.5    & \textbf{72.9}  & \textcolor{green}{\textbf{+0.4}}  \\
VoxelNet~\cite{Voxelnet}                                                                   & 69.6    & \textbf{70.6}  & \textcolor{green}{\textbf{+1.0}}  & 0.710   & \textbf{0.665} & \textcolor{green}{\textbf{-0.045}}  & 308    & \textbf{230}   & \textcolor{green}{\textbf{-78}} & 72.8    & \textbf{72.9}   &\textcolor{green}{\textbf{+0.1}} \\
ShaSTA~\cite{ShaSTA}                                                                   & 72.0    & \textbf{72.6}  & \textcolor{green}{\textbf{+0.6}}  & 0.612   & \textbf{0.593} & \textcolor{green}{\textbf{-0.019}}  & 203    & \textbf{174}  & \textcolor{green}{\textbf{-29}}  & 73.0    & \textbf{75.3}   &\textcolor{green}{\textbf{+2.3}}
\end{tabular}
\vspace{-3mm}
\label{tab: Track completion over SOTA }
\end{table*}

We first evaluate our offline tracking model on the imperfect data generated from the online tracking result. We show the relative improvements it brings to the original online tracking result on MOT metrics. In this setting, the experiments evaluate the potential of the offline tracking model to be applied in offline auto labeling. We use CenterPoint~\cite{Centerpoint} as the initial off-the-shelf detector and tracker. We also perform an intra-class non-maxima suppression with an IoU threshold of 0.1 before the initial tracking. Next, we evaluate our offline tracking model using human-annotated data. We mask the vehicle tracks to create a large amount of pseudo-occlusion cases for evaluation. The human-annotated data have little input noise, which excludes the imperfections from the online tracker so the evaluation focuses solely on the offline tracking model we proposed. Furthermore, the real occlusion cases only take a relatively small portion of the dataset compared to the unoccluded cases and the occlusion duration is typically short. Unlike the real occlusions in the dataset, the generated pseudo-occlusions are abundant and typically have longer durations. We focus our method only on the vehicle classes. All the metric scores reported are derived from the epoch in a single training run that exhibited the lowest validation error during training.
\vspace{-1mm}


\subsection{Quantitative evaluation with online tracking results}
\label{sec: online tracking eval}

\subsubsection{Re-ID model evaluation}
\label{sec: Re-ID evaluation online}
\vspace{-1mm}

In the sole evaluation of the Re-ID model, we adopt the \textbf{official nuScenes setup} which follows the standard evaluation protocol of nuScenes and filters out the empty GT boxes for evaluation. We only use the Re-ID model to reassociate the tracklets before and after occlusions, and we do not use the track completion model to interpolate the trajectories between the gaps. Instead, the nuScenes evaluation code automatically does a simple linear interpolation to fill all the gaps on both the GT tracks and the predicted tracks. For comparison, we select several top-performing lidar-based trackers that also use CenterPoint~\cite{Centerpoint} detections from the nuScenes leaderboard.

From the results in Tab.~\ref{tab: merged Re-ID AMOTA and MOT metrics test}, our model effectively improves the original online tracking result from CenterPoint~\cite{Centerpoint}. Please also note that the Re-ID model only refines the limited occlusion cases. Therefore, the improvement brought by the Re-ID model is upper bound by the number of occlusions. Yet the result after refinement already outperforms the other methods on AMOTA and AMOTP, indicating the Re-ID model can accurately reassociate the tracklets before and after occlusions. We thus have demonstrated the potential of our Re-ID model to improve the offline tracking result during auto labeling, especially for occlusion-rich scenes. 

\subsubsection{Track completion model evaluation}

Based on the Re-ID results, the track completion model recovers the missing trajectories between all the matched tracklet pairs.
Since we want to evaluate the ability of our model to recover the occluded trajectories, we adopt the \textbf{all-boxes setup} and thus do not filter out any GT bounding boxes. We take the Re-ID result refined by the motion and the map branch as input. Based on the Re-ID result, the track completion model interpolates the gaps and recovers the missing trajectories. If a gap within a track is spatially larger than 3m or temporally longer than 1.8 seconds, it is considered a possible occlusion, and the track completion model predicts the poses for the missing timestamps in between. Otherwise, the evaluation code automatically does a linear interpolation to fill the small gaps.  We show the improvements our model brings to multiple online trackers in Tab.~\ref{tab: Track completion over SOTA }. We have only tuned our model using the results from CenterPoint tracker~\cite{Centerpoint} and then applied our model to the results from other trackers. The improvements have demonstrated that our model can be used as a general plugin to recover the imperfections caused by occlusions during offline autolabeling. Our model is compatible with any initial tracker, thus it can be easily integrated into the existing offline autolabeling frameworks~\cite{offboard_labeling, Auto4d}. 

\begin{figure*}[htb]
\centering
\includegraphics[width=0.9\textwidth]{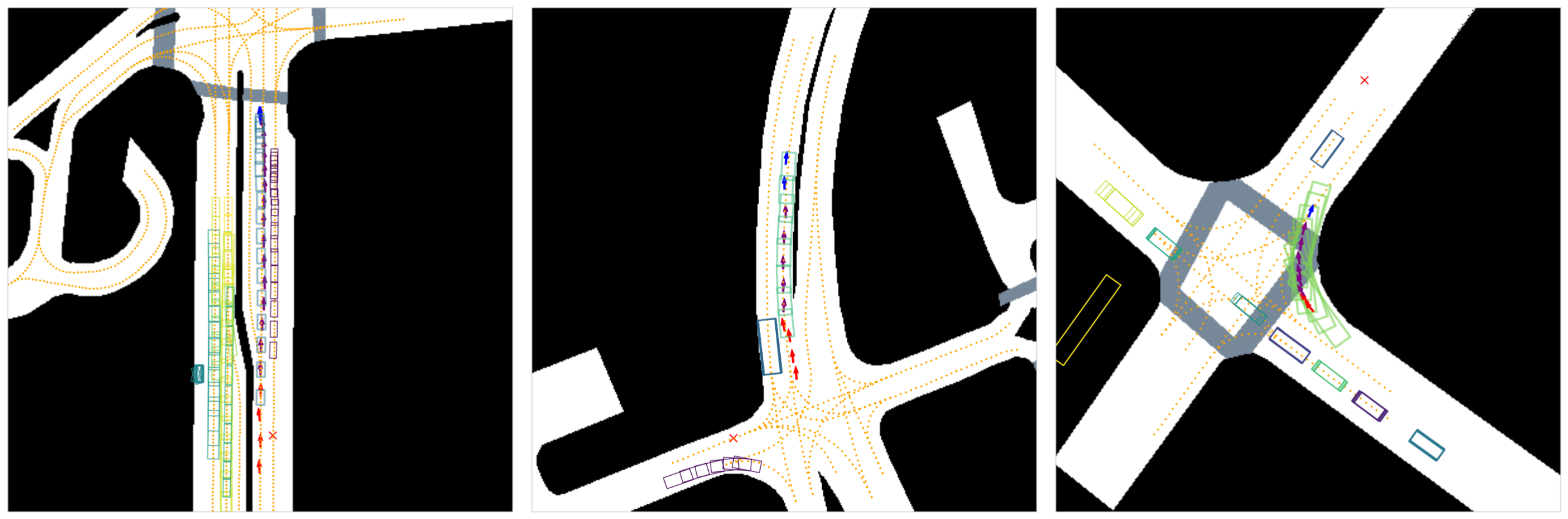}
\vspace{-2mm}
\caption{Qualitative results of the offline tracking model. In each sample, GT boxes are plotted as rectangles. Each GT track is represented by a unique color. Model outputs are depicted with arrows, rather than rectangles, to differentiate from GT tracks. Red and blue arrows indicate history and future tracklets matched by the offline Re-ID model, respectively, while purple arrows show recovered trajectories. Orange dotted lines represent lanes. In the background, the white area is drivable. The gray area is the pedestrian crossing. The red cross is the average position of the ego vehicle during the occlusion. }
\label{fig: online_tracking joint}
\vspace{-4mm}
\end{figure*}

\vspace{-1mm}
\subsection{Quantitative evaluation with pseudo-occlusions}
\label{sec: pseudo-occlusion eval}

\subsubsection{Re-ID model evaluation}
\vspace{-1mm}

\begin{table}
    \centering
    \vspace{-1mm}
    \caption{Re-ID evaluation on the nuScenes validation split (Pseudo-occlusion setup). 
    }
    \setlength\tabcolsep{5pt} 
    \begin{tabular}{c|c|c|c|c }
     \textbf{Method}  & CVM & Map & Motion & Motion+Map \\
    \hline \textbf{Association accuracy $\uparrow$} & 58.7\% & 89.6\% & 90.1\% & \textbf{90.3\%} \\
    \end{tabular}
    \vspace{-4mm}
    \label{tab: pseudo occlusion Re-ID}
\end{table}
We evaluate the Re-ID model following the \textbf{pseudo-occlusion setup}. The number of matching candidate tracklets ranges from 2 to 65. 
We also include a constant velocity model (CVM) associator as our baseline. The CVM associator takes the last observable position and velocity as inputs and then predicts future trajectories with a constant velocity. The future tracklet which has the shortest distance to the constant velocity prediction will be matched. The association accuracies are listed in Tab.~\ref{tab: pseudo occlusion Re-ID}. 
Given inputs without noise, our model in all settings achieves high association accuracies and outperforms the CVM associator by a large margin over $30\%$. With the perfect motion inputs, the motion branch has a higher association accuracy over the map branch by $0.5\%$. After the combination of the motion and the map branches, the accuracy increases by $0.2\%$ compared to the result using only the motion branch. Hence, the two branches are complementary.


\subsubsection{Track completion model evaluation}

To standardize the evaluation, for each evaluated sample track, we mask 6 seconds of its trajectory and take 2 seconds of history tracklet and future tracklet as inputs. We chose HOME~\cite{HOME} as a baseline and re-implemented it on nuScenes. HOME originally decodes the predicted trajectories from sampled endpoints. To make the comparison fair, we directly give the GT trajectory endpoints to the HOME, so that the model is also aware of the future. From the result in Tab.\ref{tab:pseudo-occ track_completion}, the performance is improved on every metric after using the map information for trajectory refinement, indicating the extracted map prior improves the trajectory recovery. Our model also outperforms HOME on both ADE and MR, demonstrating its potential to recover trajectories under long occlusions.

\begin{table}
    \centering
    \caption{Track completion evaluation on the nuScenes validation split (Pseudo-occlusion setup). 
    }
    \vspace{-1mm}
    \begin{tabular}{c|c| c| c}
    \textbf{Method} & \textbf{ADE$\downarrow$ \text{/ m}} & \textbf{Yaw error$\downarrow$ \text{/ deg}} & \textbf{MR$\downarrow$} \\
    \hline HOME~\cite{HOME} & 0.814 & - & $24.3\%$ \\
     Motion & 0.705 & 2.38 & $14.9\%$ \\
     Motion + map & $\mathbf{0 . 667}$ & $\mathbf{2.23}$ & $\mathbf{13.3\%}$ \\
    \end{tabular}
    \vspace{-4mm}
    \label{tab:pseudo-occ track_completion}
\end{table}


\vspace{-1mm}
\subsection{Qualitative results}
\vspace{-1mm}

\label{sec: Qualitative results}

As in Fig.~\ref{fig: online_tracking joint}, we applied our offline tracking model to improve the online tracking result by jointly performing Re-ID and track completion. The left plot shows a long occlusion with a duration of 8s and a gap distance of 71m. The middle plot shows a relatively short occlusion with a duration of 3s and a gap distance of 25m. The right plot shows an occlusion that happened on a turning vehicle at a crossroad. The occlusion duration in the third sample is 3.5s and the gap distance is 13m. With the three samples, we show the ability of our offline tracking model to deal with long and short occlusions, as well as the non-linear motion under occlusion. Taking the online tracking results and the semantic map as inputs, our offline tracking model first correctly associated together the tracklets belonging to the same GT track. The associated future (blue) and history (red) tracklets in each plot are covered by a single GT track. Then the model recovered the occluded trajectories by interpolating the gaps between the associated tracklet pairs. Our model is not only capable of generating accurate trajectories using lane information, but it can also infer the trajectory based on motion features when the agent deviates from surrounding lanes, as shown in the right plot. The recovered trajectories (purple) also aligned well with the corresponding GT tracks.

\section{CONCLUSION}
\vspace{-1mm}
While the previous point cloud-based offline auto labeling methods focused on generating accurate bounding boxes over visible objects, we have proposed a novel offline tracking method focusing on the occlusions on vehicles. The Re-ID module can effectively reduce IDS caused by premature termination under occlusion. Based on the Re-ID result, the track completion model recovers the occluded trajectories. Leveraging the idea from motion prediction, we innovatively extracted information from the lane map and used it as a prior to improve the performance of our models. We have demonstrated the ability of our model on the nuScenes dataset, using both the imperfect online tracking results and the handcrafted pseudo-occlusion data. The offline tracking model effectively improves the original online tracking result, showing its potential to be applied in auto labeling autonomous driving datasets. It also achieves good performances on the handcrafted pseudo-occlusions and outperforms the baselines by large margins. Future work will aim to unify the Re-ID and track completion modules into an end-to-end model and introduce a geometry affinity that utilizes cropped point cloud features to enhance association.


\vspace{-1mm}
\printbibliography[]

\clearpage

\section{APPENDIX}
\begin{figure*}[htb]
\centering
\includegraphics[width=1.0\textwidth]{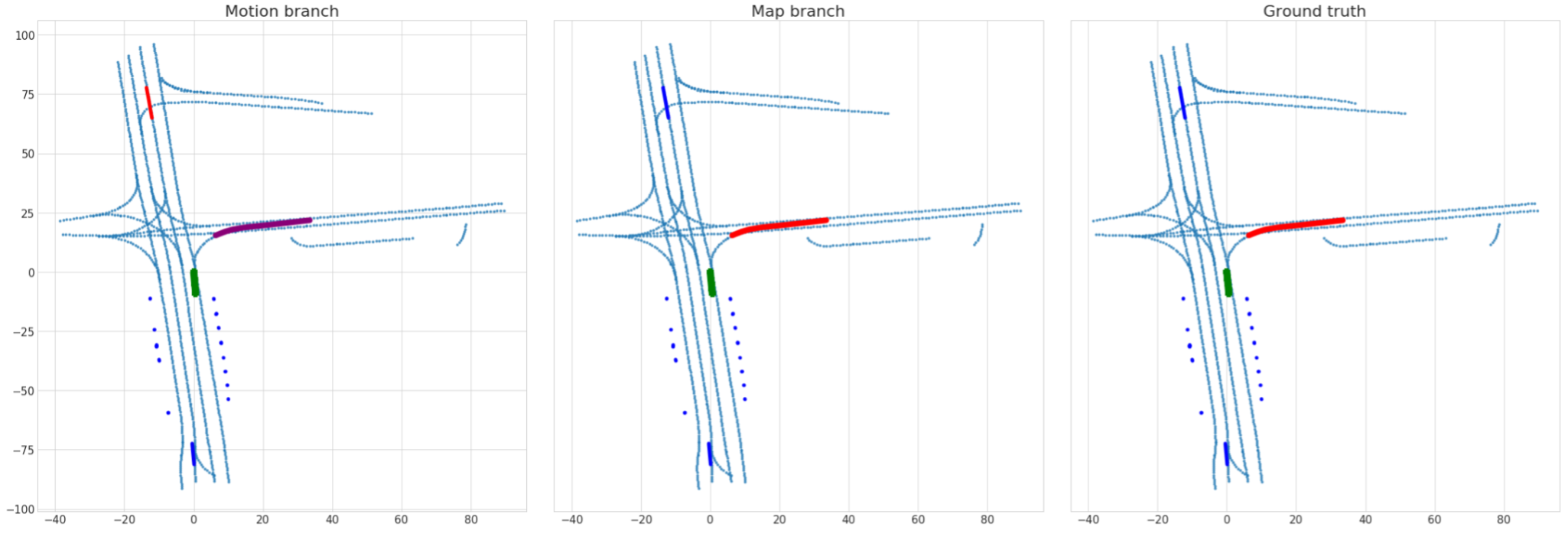}
\vspace{-2mm}
\caption{Qualitative results of the Re-ID model evaluated with pseudo-occlusion. The history tracklets are green. Future tracklets are colored according to their affinity scores. Higher scores are represented in red and lower scores are in blue. The thin dotted lines represent the lanes from the map. The left figure shows the prediction from the motion branch. The middle figure shows the prediction from the map branch, and the right figure shows the GT. }
\label{fig:pseudo-occ Re-ID}
\end{figure*}

\begin{figure*}[htb!]
\centering
\includegraphics[width=\textwidth]{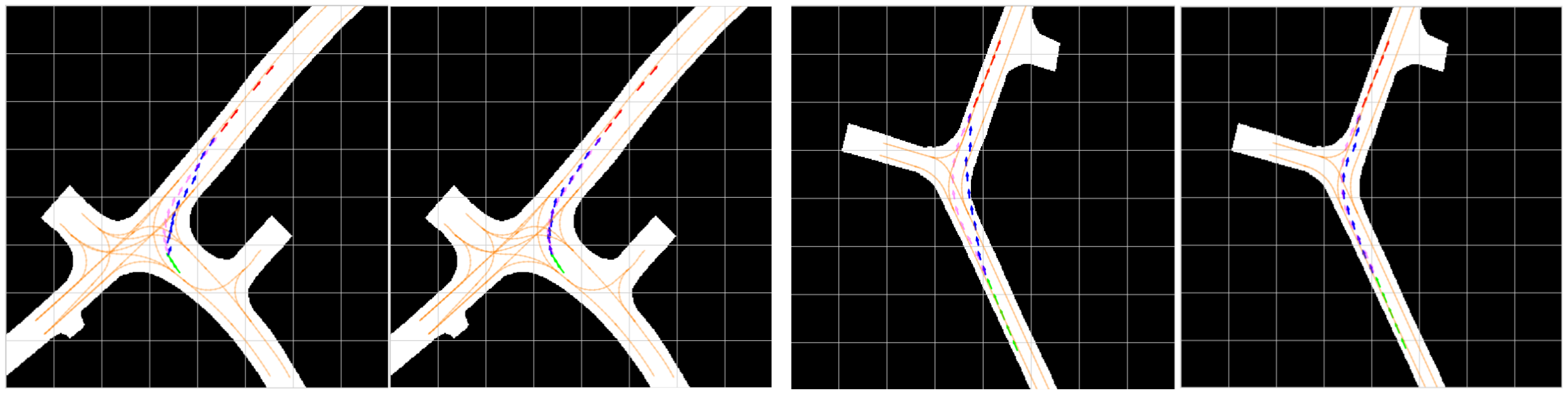}
\caption{Qualitative results of the track completion model evaluated with pseudo-occlusion. In the plot, 2 pairs of prediction results are shown. In each pair, the left image shows the decoded trajectory using only the motion feature, whereas the right image shows the trajectory after refinement. The history tracklets, future tracklets, predicted trajectories and GT trajectories correspond to the green, red, blue, and pink arrows respectively. The lanes are in orange.}
\label{fig:pseudo-occ track_completion} 
\end{figure*}

\subsection{Qualitative results on pseudo-occlusion}
\subsubsection{Re-ID model evaluation on pseudo-occlusion}

As shown in Fig.~\ref{fig:pseudo-occ Re-ID}, a representative case is selected to show the map information extracted from the lane graph can help correct the association. The history tracklets are green. Future tracklets are colored according to their affinity scores. Higher scores are represented in red and lower scores are in blue. The thin dotted lines represent the lanes from the map. The left figure shows the prediction from the motion branch. The middle figure shows the prediction from the map branch, and the right figure shows the GT. In the left figure, the motion branch tends to assign a high affinity score (red) to the future tracklet candidate above. Though this association fits the motion pattern, it is a false match. After adding the map information, the affinity score of the false future candidate is suppressed, since it is not on the same lane as the history tracklet. The affinity score of the true future tracklet increases in the map branch, showing that our offline tracking model can utilize the map prior to accurately associating tracklets.

\subsubsection{Track completion model evaluation on pseudo-occlusion}

In Fig.~\ref{fig:pseudo-occ track_completion}, the left image shows the decoded trajectory using only the motion feature, whereas the right image shows the trajectory after refinement. The history tracklets, future tracklets, predicted trajectories and GT trajectories correspond to the green, red, blue, and pink arrows respectively. The lanes are in orange. After refinement, the trajectories are smoother and tend to follow the lane. Less deviation from the GT trajectory is observed. Hence it accords with our previous quantitative evaluation, showing our track completion model is able to leverage the map prior to generating accurate trajectories.

\subsection{Loss functions} 
\label{sec: loss formulas}
We adopt the focal loss~\cite{focal_loss} to train the Re-ID model.

\begin{equation}
\label{eq: focal loss}
\begin{aligned}
\mathcal{L}_{cls}\left(k_{\mathrm{t}}\right)=-\alpha_{\mathrm{t}}\left(1-k_{\mathrm{t}}\right)^\gamma \log \left(k_{\mathrm{t}}\right) \\
k_{\mathrm{t}}= \begin{cases}k & \text { if } y=1 \\ 1-k & \text { otherwise }\end{cases}
\end{aligned}
\end{equation}
where $k$ is the predicted affinity score, and $y$ is the GT score. $\alpha_{\mathrm{t}}$ and $\gamma$ are two hyperparameters, which are set to $0.5$ and $2.0$ respectively. 

For the track completion model, we use Huber loss (i.e. smooth L1 loss) to train the $x,y$ coordinate regression heads in both the initial trajectory decoder and the final refinement blocks: 
\begin{equation}
\begin{aligned}
\mathcal{L}_{\text {coord }} & =\frac{1}{t_f} \sum_{t=1}^{t_f} d\left(m_t-m_t^{g t}\right) \\
d\left(x_i\right) & = \begin{cases}0.5\left\|x_i\right\|_2 & \text { if }\left\|x_i\right\|<1 \\
\left\|x_i\right\|-0.5 & \text { otherwise }\end{cases}
\end{aligned}
\end{equation}
where $m_t = \left[x_t,y_t\right]$ are the predicted $x,y$ coordinates in local agent frame, $t_f$ the prediction horizon. $\left\| \cdot \right\|$ and $\left\|\cdot \right\|_2$ denotes the $\ell_1$ and $\ell_2$ norm . 
For yaw angle regression, we first adjust all the GT yaw angles such that their absolute differences with the predicted yaw angles are less than $\pi$. 
Then, we apply $\ell_1$ loss on the regressed yaw angles:
\begin{equation}
\label{eq: yaw regression loss}
\begin{aligned}
\mathcal{L}_{\text {yaw }}=\frac{1}{t_f} \sum_{t=1}^{t_f}\left|\theta_t-\theta_t^{g t}\right|
\end{aligned}
\end{equation}
The final regression loss $\mathcal{L}_{\text {reg }}$ is:
\begin{equation}
\label{eq: total regression loss}
\begin{aligned}
\mathcal{L}_{\text {reg }}=\alpha_{coord}\mathcal{L}_{\text {coord }} + \alpha_{yaw}\mathcal{L}_{\text {yaw }}
\end{aligned}
\end{equation}
where $\alpha_{coord}$ and $\alpha_{yaw}$ are the weights for the coordinate regression loss $\mathcal{L}_{\text {coord }}$ and yaw angle regression loss $\mathcal{L}_{\text {yaw }}$ respectively. We set $\alpha_{coord}=1.0$ and $\alpha_{yaw}=0.5$.

\subsection{Implementation details}
\label{sec: implementation details}
We train the Re-ID model for 50 epochs on Tesla V-100 GPU using a batch size of 64 with the AdamW optimizer with an initial learning rate of $1 \times 10^{-3}$, which decays by a factor of 0.6 every 10 epochs. The training time is 11.5 hours. We train the track completion model following the same setting except for a different decay factor of 0.5 every 10 epochs. The training time is 4.2 hours. 

Our model is agnostic to the selections of the detector and online tracker. For the evaluation with online tracking results, we use an association threshold of 0.9. If the the motion affinity and map-based affinity scores are both lower than the threshold, the corresponding matching pair would not participate in the association. For the evaluation with pseudo-occlusions, we use the tracks from the nuScenes validation split provided by the nuScenes software devkit for the motion prediction challenge. 
\subsection{Supplement of experimental setups}
\label{sec: experimental setup}
\subsubsection{Experimental setup for track completion model evaluation with online tracking results}
\label{sec: Track completion evaluation online setup}
We take the Re-ID result refined by the motion and the map branch as input. Based on the Re-ID result, the track completion model interpolates the gaps and recovers the missing trajectories. If a gap within a track is spatially larger than 3m or temporally longer than 1.8 seconds, it is considered a possible occlusion, and the track completion model predicts the poses for the missing timestamps in between. Otherwise, the evaluation code automatically does a linear interpolation to fill the small gaps. The sizes of the recovered bounding boxes are linearly interpolated from the two bounding boxes in the history tracklet and the future tracklet. 
\subsubsection{Experimental setup for Re-ID model evaluation with pseudo-occlusions}
\label{sec: Re-ID evaluation pseudo-occ setup}
\begin{figure*}[htb]
   \begin{minipage}{0.48\textwidth}
     \centering
     \includegraphics[width=\textwidth]{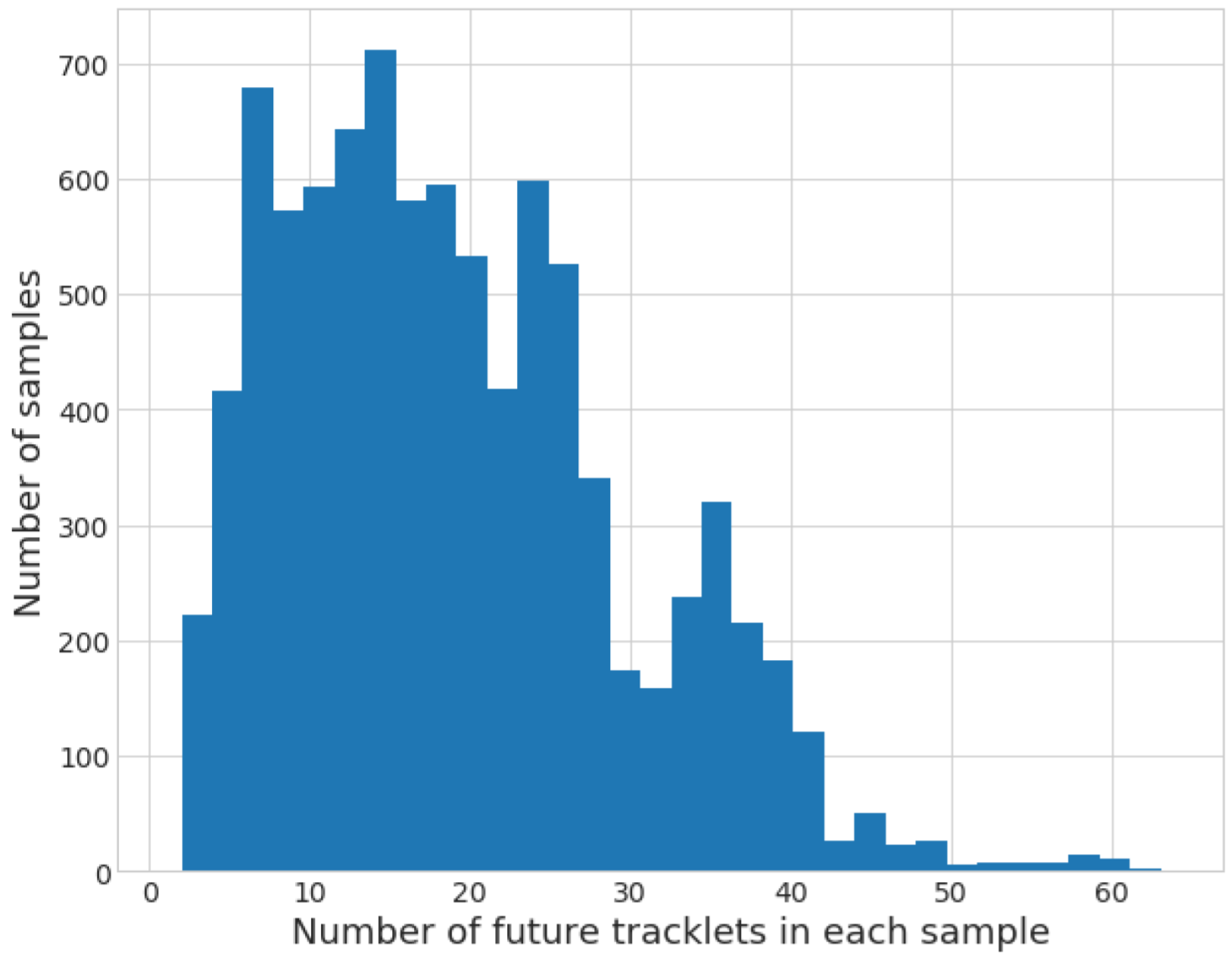}
    \caption{The distribution of the number of candidate tracklets in Re-ID model evaluation with pseudo-occlusions. }
    \label{fig: track_num_distribution}
   \end{minipage}\hfill
   \begin{minipage}{0.48\textwidth}
    \centering
    \includegraphics[width=\textwidth]{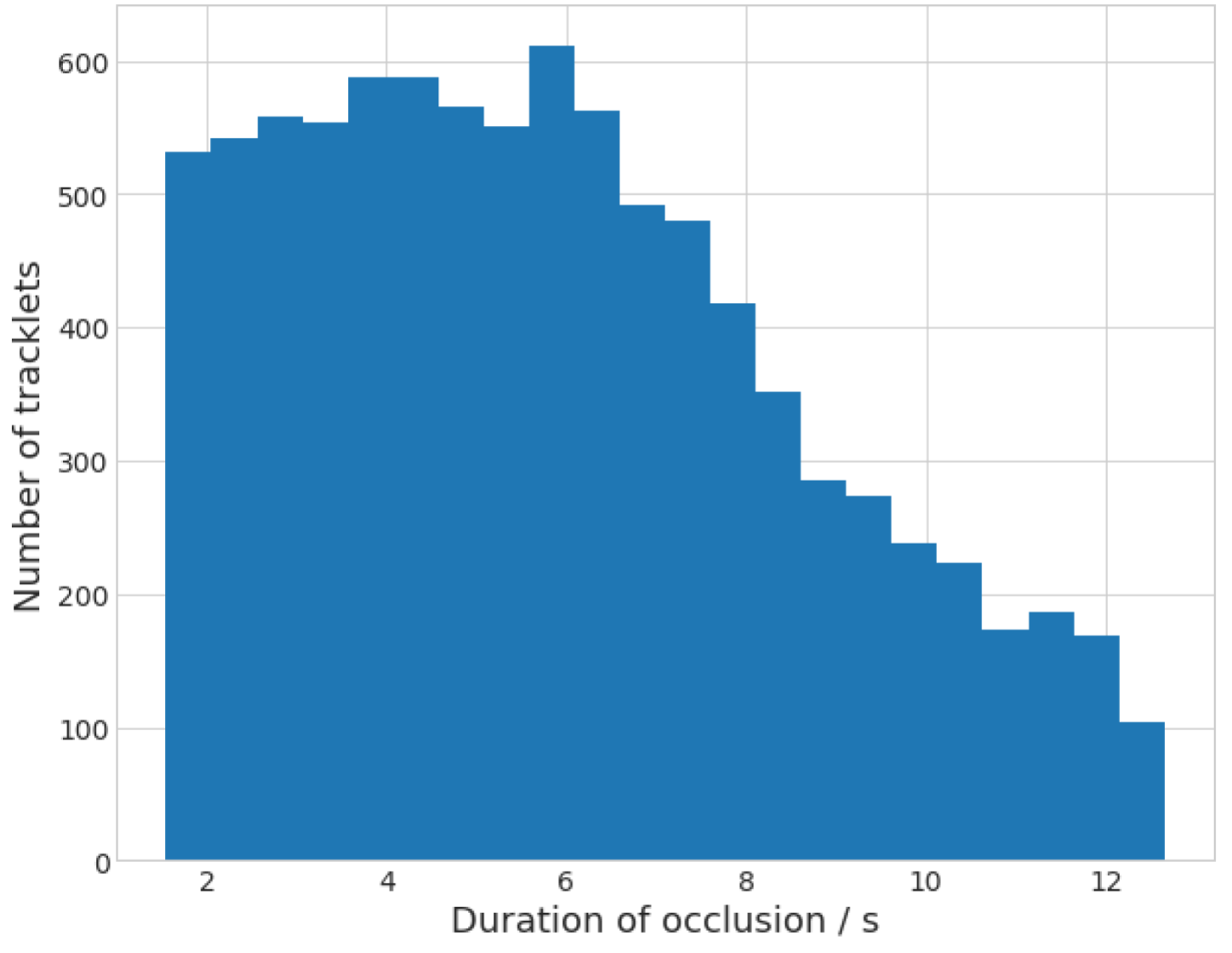}
    \caption{The distribution of the pseudo occlusion durations in Re-ID model evaluation. }
    \label{fig: psd_occ_length_distribution}
   \end{minipage}
\end{figure*}

We mask the GT tracks in the nuScenes validation split to create pseudo-occlusions to evaluate our Re-ID model independently. In every evaluated sample, all future tracks are masked for random durations to create diverse pseudo-occlusions. The durations range from 1.5s to each track's maximal length (12.5s at most) such that each future tracklet has at least one last pose visible. The history tracklet in each sample is also randomly deprecated such that it has at least one pose left as input and is at most 2.5 seconds long. The number of matching candidate tracklets ranges from 2 to 65, as shown in Fig.~\ref{fig: track_num_distribution}. The distribution of pseudo-occlusions is shown in Fig.~\ref{fig: psd_occ_length_distribution}. 

\end{document}